\documentclass[journal,twoside,web]{ieeecolor_nologo}
\providecommand{\Description}[1]{}
\newcommand{\sig}[1]{\textsuperscript{#1}}
\usepackage{float}
\usepackage{generic_nologo}
\usepackage{cite}
\usepackage{amsmath,amssymb,amsfonts}
\usepackage{graphicx}
\usepackage{algorithm,algorithmic}
\usepackage{hyperref}
 
\hypersetup{hidelinks}
\usepackage{textcomp}

\usepackage{makecell}
\def\BibTeX{{\rm B\kern-.05em{\sc i\kern-.025em b}\kern-.08em
    T\kern-.1667em\lower.7ex\hbox{E}\kern-.125emX}}
\usepackage{stfloats}
\let\appendices\relax
\usepackage[title,toc]{appendix}
\usepackage{cuted}
\usepackage{fancyhdr}
\usepackage{multirow}
\usepackage{booktabs}
\usepackage{silence}
\usepackage{caption}
\usepackage{subcaption}
\newtheorem{hyp}{Hypothesis}

\begin{document}
\title{HealthCAT: An Interpretable Encoder-only Transformer Framework for Health Indicator Prediction and Temporal Interpretation of Wearable Sensor Data}

\author{Xiaotong Yu, Joshua Y. Kim, HaeJin Lee, Kalina Yacef.
\thanks{This manuscript is submitted to Journal of Biomedical and Health Informatics on 1 June 2026.}
\thanks{
Xiaotong Yu is with School of Computer Science, The University of Sydney, Sydney, Australia, NSW 2008 (e-mail: xiaotong.yu@sydney.edu.au). }
\thanks{
Joshua Y. Kim is with School of Computer Science, The University of Sydney, Sydney, Australia, NSW 2008 (e-mail: josh.kim@sydney.edu.au).}
\thanks{
HaeJin Lee is with School of Computer Science, The University of Sydney, Sydney, Australia, NSW 2008 (e-mail: haejin.lee@sydney.edu.au).}
\thanks{
Kalina Yacef is with School of Computer Science, The University of Sydney, Sydney, Australia, NSW 2008 (e-mail: kalina.yacef@sydney.edu.au)}}

\maketitle

\begin{abstract}
Wearable sensors continuously capture fine-grained multivariate time-series data, providing opportunities to model behavioural patterns associated with health outcomes. However, existing deep learning methods prioritise predictive accuracy over interpretability, limiting their application in health research. In this study, we present HealthCAT, a flexible framework that integrates an Encoder-only Transformer with an Attentive Class Activation Token (AttentiveCAT) to generate class-specific, time-step-level interpretations. These interpretations can be mapped back onto behavioural cycles that are relevant to the domain (e.g., time-of-day), supporting individual-level analysis of wearable sensor data. We evaluated HealthCAT using two real-world wearable sensor datasets (306 participants in total). HealthCAT outperformed deep learning baselines by up to 17\% in F1-score and 12\% in accuracy on both datasets ($p<0.05$). In masking experiments, the time steps identified by HealthCAT carried significantly more predictive value than random selection across all masking conditions ($p<0.05$), indicating that the identified time steps are predictively informative. By coupling predictive performance with validated time-step-level interpretability, HealthCAT moves wearable sensor analysis beyond aggregated metrics towards temporal patterns that support health monitoring, behavioural pattern analysis, and intervention design in health research. The significance of this work is that it enables accurate prediction of health indicators from wearable sensor data while providing insights into when and how physical activity patterns occur, rather than relying solely on aggregated summary measures.
\end{abstract}


\section{Introduction}
\label{sec:introduction}
From commercial fitness trackers to research-level wearables, sensors are now embedded in millions of devices worldwide, continuously collecting movement and physiological signals \cite{ravi2016deep, khurshid2022wearable}. These sensors can be used to monitor physical activity, sleep patterns, and related health features in a real-world environment \cite{henriksen2018using}, offering opportunities to study user behaviours and their associations with health outcomes, such as body weight, sleep quality, and chronic conditions \cite{braem2025association, moshawrab2023smart,ahmed2022overview,lubitz2022detection}. However, the rich sensor data have been reduced to aggregated metrics, such as daily total step counts or daily total moderate-to-vigorous physical activity (MVPA) \cite{ferguson2022effectiveness, jorke2025gptcoach, WHO_BeActive}. These metrics are valuable and serve as the gold standard in health research, but they may overlook when and how temporal patterns impact health outcomes. Therefore, moving beyond daily summaries to time-aware behavioural modelling helps to discover meaningful patterns and motivates the application of deep learning methods to capture these patterns. However, while existing approaches have demonstrated the potential of deep learning for wearable sensor data \cite{ravi2016deep, phan2021xsleepnet, perslev2021u, raza5021631lightweight, khan2022capslstm,10064701}, few studies provide fine-grained interpretations that highlight when specific behaviours impact health outcomes \cite{verde2022wearable}. For instance, deep learning has enhanced recognition tasks on wearable data, such as Human Activity Recognition (HAR) and sleep staging, by learning directly from processed raw signals \cite{miao2024spatial, zhou2024autoaughar}. However, in relation to health outcome prediction tasks, recognising which activity has occurred is not sufficient. Health and behavioural research also requires evidence of when activities occurred and at what intensity level, as well as whether such temporal evidence is stable enough to support feedback or intervention. Without such evidence, the integration of deep learning models into health research remains constrained, even though their predictive performance is reliable.

Wearable sensors provide continuous data streams, but the behavioural patterns relevant to health outcomes are typically interpreted as part of everyday routines (e.g., school/work schedules, sleep cycles). This leaves a gap between aggregated features (e.g., daily average steps, total weekly sleeping time) and temporal features for health monitoring. To address this gap, we introduce HealthCAT, a flexible Encoder-only Transformer framework with the Attentive Class Activation Token (AttentiveCAT) \cite{zerveas2021transformer,qiang2022attcat}, which generates importance scores for each time step that contribute to the predictions. In this study, the term 'time step' refers to the discrete time points in the multivariate time series, and its interval depends on the data granularity at which the wearable sensor features are processed for each dataset (Section~\ref{sec:method}) \cite{zerveas2021transformer}. These time step importance scores will be further domain-adapted for post hoc analysis, allowing the direct examination of the contributions of specific activity patterns to a prediction (e.g., when does behaviour matter?). We evaluate HealthCAT from two aspects: one aspect is predictive performance, and the other is its time-step-level interpretability. The evaluation was conducted on two real-world wearable sensor datasets and two health indicators: (1) a physical activity dataset of 206 Pacific Island adolescents, predicting their healthy weight status \cite{wattelez2025assessing}; (2) a sleep dataset of 100 adults sleep status and time aligned multisensor data for identifying temporal sleep patterns linked to sleep health \cite{wang2024dreamt}. 

To our knowledge, HealthCAT is the first study to provide a time-step-level interpretation in health wearable sensor data modelling. Our contributions are listed as follows:

\begin{itemize}
    \item \textbf{Interpretable Framework.} We propose HealthCAT, a flexible Encoder-only Transformer framework that integrates AttentiveCAT to produce class-specific, time-step-level interpretations from multivariate wearable sensor data.
    
    \item \textbf{Interpretability Evaluation.} We evaluate HealthCAT's time step-level interpretation, confirming that the identified time steps carry more predictive power than random selection.
    
    \item \textbf{Domain-Adapted Visualisation.} We provide a domain-adapted way to visualise HealthCAT's interpretations at the individual level.
    
\end{itemize}

The source code will be published on GitHub. 

\section{RELATED WORKS}
\label{sec:relatedwork}
\subsection{Deep Learning for Wearable Sensor Data}
Deep learning models have demonstrated outstanding performance in time series analysis \cite{singh2020deep, liu2021gated, wen2022transformers, le2024shapeformer}, such as Long-Short Term Memory (LSTM) \cite{graves2012long}, Gated Recurrent Unit (GRU) \cite{chung2014empirical}, and Transformer \cite{vaswani2017attention}. The variations of deep learning models are commonly used for modelling wearable sensor data in the health and medical domains across different downstream tasks, such as human activity recognition \cite{raza5021631lightweight, khan2022capslstm,10064701}, sleep staging \cite{phan2021xsleepnet, perslev2021u}, and health outcomes screening \cite{braem2025association, moshawrab2023smart,lubitz2022detection}. For example, DeepSleepNet combines Convolutional Neural Network (CNN) and Bi-directional LSTM to categorise sleep stages using electroencephalogram (EEG), achieving good performance with limited focus on explanations \cite{supratak2017deepsleepnet}. Similarly, \cite{rajpurkar2017cardiologist} proposed a CNN-based method for detecting arrhythmias from an Electrocardiogram (ECG), demonstrating strong predictive power but still lacking interpretability in its predictions. There are also Transformer-based methods to detect activity using accelerometer and Photoplethysmography (PPG) data \cite{sharma2022transformer} without an interpretability layer. These studies have performed well in downstream tasks, but most have focused on signal data; few have paid attention to time series data from wearable sensors. Recent work has moved towards label-efficient learning on wearables \cite{yu2023semi} and compact Encoder-only Transformer for multivariate time series \cite{zerveas2021transformer, wen2022transformers,lubitz2022detection, le2024shapeformer}. However, these approaches typically optimise accuracy while offering limited explanations.

Early deep learning work on wearables primarily focused on HAR and sleep staging tasks using annotated datasets and consistent labels, such as walking, swimming, Rapid Eye Movement, and Non-Rapid Eye Movement \cite{miao2024spatial,khan2022capslstm,perslev2021u,phan2021xsleepnet,raza5021631lightweight}, establishing that deep learning models learn discriminative representations directly from raw or minimally processed accelerometer and PPG \cite{raza5021631lightweight,sharma2022transformer,singh2020deep,chung2014empirical}. More recently, research has begun to target health outcomes, such as stress, cardiovascular disease risk, or sleep disordered breathing, where ground truth is sparse and behavioural outcome links are mediated by context and time (e.g., work/school routines). The lack of interpretability is therefore the primary challenge in integrating deep learning methods into health research, where experts require interpretability to validate the data-driven findings \cite{anand2026self, verde2022wearable}. An interpretable model is not just desired but necessary to translate model outputs into domain-adapted insights.

\subsection{Interpretability Methods}

While state-of-the-art deep learning models capture complex temporal dependencies in wearable sensor data, they often provide limited explanation of which time periods contribute to a specific prediction \cite{garcia2024sleepecg, abdelaal2024exploring}. For wearable health analytics, effective interpretability should consider both the model's decision boundary and the temporal granularity (e.g., minutes or hours) \cite{rossi2023sleep}. Several post hoc methods have been developed to interpret model behaviour: self-attention interpretation \cite{li2019enhancing, phan2022sleeptransformer}, GradCAM \cite{selvaraju2017grad}, and model-agnostic tools such as DeepSHAP \cite{lundberg2017unified} and LIME \cite{ribeiro2016should}. Each of these methods has limitations for time-step-level interpretation of wearable sensor data. For instance, attention visualisation maps weights to inputs, but it is not class-specific and is not guaranteed to reflect the decision boundary, which limits its reliability for explaining predictions \cite{jin2024cardiologist, phan2022sleeptransformer}. \cite{jin2024cardiologist} overlaid spatial attention on ECG recordings for arrhythmia detection and qualitatively evaluated the attention maps. Another attention-based interpretation was proposed in SleepTransformer \cite{phan2022sleeptransformer}, which visualised attention scores at both epoch and sequence levels; however, the scores are not class-specific and are tied to fixed 30-second epochs. GradCAM is designed for CNN-based encoders, providing discriminative visualisations to explain predictions. However, the map-based explanation does not transfer naturally to sequential models \cite{selvaraju2017grad, barkan2021grad}. Similarly, DeepSHAP can be applied to time series but is computationally expensive for long sequences and does not always preserve temporal dependencies. Its attributions are typically feature-level rather than localised to specific time steps. As another method, \cite{climeforts} adapted LIME to explain high-level concepts in time-series predictions, but not at time-step-level granularity. Generic post-hoc tools (LIME, SHAP, GradCAM) provide useful feature-attribution perspectives \cite{climeforts, lundberg2017unified, ribeiro2016should, selvaraju2017grad, barkan2021grad}, yet they are not optimised for long, multi-sensor sequences, where \emph{when} a pattern occurs is as important as \emph{what} occurs.

Class activation mechanisms address this gap by combining a class score with internal representations to yield class-specific contributions to the model's decision \cite{selvaraju2017grad}. The Attentive Class Activation Token (AttentiveCAT) extends this idea to Transformer by integrating class activation with self-attention to produce time-step-level importance scores \cite{qiang2022attcat}. Originally developed to explain pre-trained NLP Transformer, AttentiveCAT can be adapted to multivariate wearable time series by treating each time step analogously to a token, yielding class-specific importance scores that can be projected back onto the original time axes.

The studies reviewed above highlight three key gaps in the application of deep learning to wearable sensor data in health research: (1) the lack of a framework that captures temporal patterns with interpretability; (2) the absence of evaluation method for validating temporal interpretation; (3) limited approaches for mapping these temporal interpretations onto domain-relevant cycles. 
Motivated by these gaps, our work addresses the following research questions:

\textbf{RQ1}: Does HealthCAT improve predictive performance over baselines for predicting specific health indicators from wearable sensor data?

\textbf{RQ2}: Do the time steps identified by HealthCAT have more predictive information than random selection? 

\textbf{RQ3}: Can we surface the behavioural patterns in order to facilitate the interpretability of time-step-level importance scores meaningfully for the given domain?

\section{METHODOLOGY}\label{sec:method}
\begin{figure*}[htbp]
    \centering
    \includegraphics[width=\textwidth]{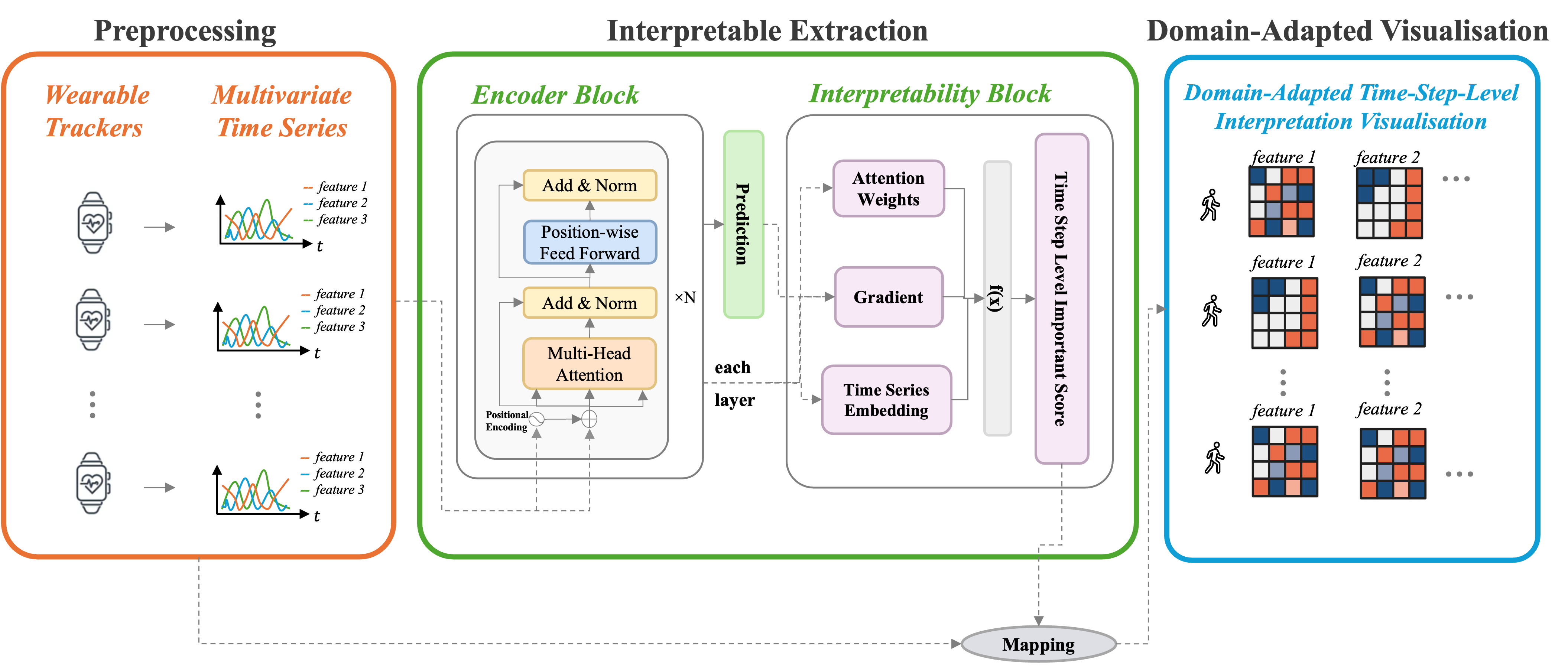}
    \caption{Overview of HealthCAT framework}
    \Description{}
    \label{fig:overview_method}
\end{figure*}

The HealthCAT framework is a three-stage framework for predicting health indicators from wearable sensor data while providing time-step-level interpretability. In the first stage, preprocessing, raw wearable device data are converted into a structured multivariate time series representation suitable for deep learning. In the second stage, interpretable extraction, an Encoder-only Transformer model is trained to predict health outcomes, and the time-step-level importance scores are computed from the model's outputs. In the third stage, domain-adapted visualisation, the time-step-level importance scores are mapped back onto the original multivariate time series to produce individual-level visualisations aligned with domain-relevant behavioural cycles. An overview of HealthCAT framework is shown in Fig.~\ref{fig:overview_method}.

\subsection{Datasets}\label{sec:dataset}
We included two real-world wearable sensor datasets from the health domain to evaluate the proposed framework: (1) physical activity dataset \cite{wattelez2025assessing} and (2) sleep dataset named DREAMT \cite{wang2024dreamt}, both collected via wearable devices, as summarised in ~Table~\ref{dataset_stat}.

\textbf{Physical Activity dataset} contains anonymised data from 206 adolescents aged 10-to-16 years in the Pacific Islands, collected by \cite{wattelez2025assessing} between July 2018 and April 2019. Physical activity data were recorded continuously over 7 days using the wrist-worn GENEActiv device \cite{Dodge_2025}, which features 60 Hz triaxial accelerometers in a free-living environment. The dataset also includes age, gender, Tanner stage (also known as sexual maturity rating), and body mass index (BMI). Following \cite{hfz_category}, we selected Healthy Weight Status (HWS) as the target health indicator. HWS is an evaluation metric of an adolescent's fitness performance, calculated based on an individual's age, gender, and BMI. The original four HWS categories were merged into binary prediction task: Very Lean into Healthy Fitness Zone (including Very Lean) and Need Improvement (including Need Improvement High Risk). The label distribution was  74.3\% Healthy Fitness Zone and 25.7\% Need Improvement. 

\textbf{DREAMT dataset} contains anonymised data from 100 adults aged 21 to 86 years, collected by Duke University Health System Sleep Disorder Lab between May 2022 and September 2022 \cite{wang2024dreamt}. Data were collected using the Empatica E4 wristband \cite{EmpaticaE4}, a multi-sensor embedded device, from approximately 8:00 PM to 6:00 AM the following morning. The device captures blood volume pulse, triaxial acceleration, electrodermal activity, and skin temperature; heart rate and inter-beat interval are derived from blood volume pulse. All signals are resampled to 100 Hz and synchronised using timestamps. Following the dataset recommendation, we used the Apnea-Hypopnea Index (AHI) as the target sleep health indicator \cite{wang2024dreamt}. Based on clinical guidance \cite{american1999sleep}, AHI was converted into a binary prediction task: healthyAHI ($<5$/h) and unhealthyAHI ($ >=5$/h). The label distribution was 74\% healthyAHI and 26\% unhealthyAHI. 

\begin{table}[ht]
    \centering
    \caption{Dataset Statistics}
    \label{tab:dataset_stat}
    \scriptsize
    \setlength{\tabcolsep}{2pt}
    \renewcommand{\arraystretch}{1.1}
    
    \begin{tabular}{@{}l c c l l c@{}}
        \toprule
        Dataset & $N$ & Duration & Device & Placement & Rate \\
        \midrule
        PA \cite{wattelez2025assessing} 
        & 206 & 7 d & GENEActiv \cite{Dodge_2025} 
        & Non-dom. wrist & 60 Hz \\
        
        DREAMT \cite{wang2024dreamt} 
        & 100 & 6--8 h & Empatica E4 \cite{EmpaticaE4} 
        & Left wrist & 100 Hz \\
        \bottomrule
    \end{tabular}
    
    \vspace{2pt}
    \footnotesize{\textit{Note.} PA:Physical Activity; $N$:number of samples; Non-dom.:non-dominant.}
\label{dataset_stat}
\end{table}

\subsection{Multivariate Time Series Representation}
Wearable devices in the health domain can integrate multiple sensors, including triaxial accelerometers, PPG, and skin temperature sensors, to collect raw physiological and movement data. To process these multi-sensor signals, we designed a hybrid preprocessing method to generate multivariate time series representations for the prediction model, as shown in Fig.~\ref{fig:overview_method}. 

For processing accelerometer data, when the focus is on capturing physical activity, we compute the gravity-subtracted Signal Vector Magnitude (SVMgs), as defined in ~Equation~\ref{eq:svm_gs}, and categorise intensity levels based on previously established thresholds \cite{Phillips2013}. The other option is to compute the Standard Signal Vector Magnitude (SVM) ~\ref{eq:svm} when the gravity component is not required for movement analysis, such as in sleep data. The health-related features extracted from other sensors are concatenated with the preprocessed accelerometer data. Since wearable sensor data were recorded at varying sampling frequencies, we aggregated these features according to predefined time step units. For example, physical activity data were aggregated in hourly intervals, whereas sleep data were aggregated in minute intervals. In the end, we obtained a structured multivariate time series representation with consistent time steps for each participant.

We applied the preprocessing method to the two datasets stated in the previous section. \textbf{Physical Activity dataset} was collected using a tri-axial accelerometer only; therefore,the raw accelerometer signals were converted into SVMgs values for each second. A fixed bout length threshold has been set to categorise activity intensity levels. The bout length threshold is 3 seconds for MVPA and Light Physical Activity (LPA), and 60 seconds for Sedentary (SED). For example, if MVPA continued to occur for 18 seconds, it would be counted as 6 bouts. In this case, the number of bouts is equal to 6, and the total time spent on MVPA is equal to 18 seconds. Using established thresholds by \cite{phillips2013calibration}, we categorised into physical activity intensity levels into MVPA, LPA, as well as SED. The final time series representation is at an hourly interval, where we recorded the number of bouts and the total seconds spent in each hour for each intensity level. \textbf{DREAMT dataset} was collected via multiple sensors. The accelerometer data were computed via SVM to extract an accelerometer\_magnitude feature and then concatenated with data from other sensors. As the DREAMT data were recorded for approximately 10 hours, they have been resampled to a standard frequency of 100 Hz. Therefore, we processed the DREAMT data into minute-level resolution, recording the accelerometer\_magnitude, blood volume pulse, inter-beat interval, electrodermal activity, skin temperature, heart rate, and sleep stage at each time step.

\begin{equation}
\text{SVM}_{gs} = \sum_{i=1}^{n} \left| \sqrt{x_i^2 + y_i^2 + z_i^2} - g \right|
\label{eq:svm_gs}
\end{equation}

\noindent where $x_i, y_i,z_i$ represents raw acceleration values from tri-axial orthogonal axes ($X, Y, Z$) at the $i$-th sample, $n$ represents the number of accelerometer samples within the chosen time window, and $g$ represents standard gravitational acceleration.

\begin{equation}
\text{SVM} = \sqrt{x_i^2 + y_i^2 + z_i^2}
\label{eq:svm}
\end{equation}

\noindent where $x_i, y_i,z_i$ is as stated above.

\subsection{Interpretable Extraction}\label{sec:method_rq2}
We applied the Encoder-only Transformer for multivariate time series \cite{zerveas2021transformer} to capture the temporal dynamics in wearable sensor data, and we integrated the Attentive Class Activation Token (AttentiveCAT) \cite{qiang2022attcat} for time-step-level interpretation. 

Each training sample $X \in \mathbb{R}^{w \times m} = [x_1, x_2, ..., x_w]$ represents a multivariate time series of length $w$ with $m$ variables, containing a sequence of $w$ feature vectors $x_t \in \mathbb{R}^{m}$ at time step $t$. Each time step vector $x_t$ contains the preprocessed wearable sensor features for that time segment. All inputs were normalised over the training set and then projected into a $d$-dimensional vector space, producing a sequence $\{u_1, u_2, \dots, u_w\}$ for Transformer processing. The Encoder-only Transformer contained $L$ stacked encoder layers with learnable positional encoding to capture temporal dependencies. Each layer applied multi-head self-attention, followed by a feed-forward network, both of which employed residual connections and layer normalisation. The attention mechanism computes pairwise attention weights $\alpha_{i,j}^l$ between time steps $i$ and $j$ on layer $l$. We averaged the attention weights over all heads for each layer to simplify the calculation of AttentiveCAT scores while preserving overall pairwise relationships between time steps. The hidden representation for time step $t$ at layer $l$ is denoted as $h_t^l \in \mathbb{R}^{d}$; representations for every layer are retained for the AttentiveCAT computation below. The final layer representations are concatenated as $\bar{h} \in \mathbb{R}^{d \cdot w}= [h_1^L; \dots; h_w^L] $ and passed through a linear classification layer that computes the pre-softmax class logit $y^c$ for class $c$. For a specific class $c$, we extracted the classification weight $W_{o,t}^c$ associated with time step $t$. 

Firstly, to compute the Class Activation Token (CAT) \cite{qiang2022attcat, selvaraju2017grad} for time step $t$, we took the element-wise product between hidden representation $h_t^l$ and the gradient of the class logit $y^c$ with respect to $h_t^l$. At the final layer, the gradient is equivalent to $W_{o,t}^c$. CAT therefore reflects how the representation of time step $t$ contributes to the classification decision for class $c$:
\begin{equation}
    \mathrm{CAT}_t^{c,l} = \nabla_{h_t^l} y^c \odot h_t^l
\end{equation}

where $\odot$ represents for element-wise product. 
Secondly, we incorporated contextual relationships by combining the CAT with head-averaged self-attention weights $\alpha_t^l$ at each layer $l$. This attention-weigted CAT is referred as AttentiveCAT score:
\begin{equation}
    \mathrm{AttentiveCAT}_t^{c,l} = \alpha_t^l \cdot \mathrm{CAT}_t^{c,l}
\end{equation}

After summing the AttentiveCAT score across all layers and reducing over the hidden dimension, we obtained the final class-specific score $s_t^c$ for each time step:
\begin{equation}
    s_t^c = \sum_{l=1}^L \mathrm{AttentiveCAT}_t^{c,l}
\end{equation}

The score sequence $S^c = [s_1^c, s_2^c, ...,s_w^c]$ is used in the domain-adapted visualisation step described in Section~\ref{sec:method_rq3}.

\subsection{Domain-Adapted Visualisation}\label{sec:method_rq3}
The interpretation extraction step in Fig.~\ref{fig:overview_method} produces class-specific AttentiveCAT scores $S^c = [s_1^c, s_2^c, ...,s_w^c]$. However, these scores are originally indexed by model input sequence (i.e., the multivariate time series) and are not directly corresponding to how domain experts inspect wearable health data. Therefore, we further map the AttentiveCAT scores $S^c$ back onto domain-relevant behavioural cycles and feature dimensions. 
We use $\rho(t)$ to map each valid time step $t$ to a domain-relevant temporal position $k$, such as hour of day, minute of sleep recording. We define a domain feature function $\phi_r(x_t)$ that extracts the value of a selected domain feature $r$ from the original input data. Depending on the application, this feature can represent physical activity intensity level, sleep-related physiological measures, etc. If the feature values need to be grouped into interpretable ranges, we offer the bucket function $\operatorname{bucket}_r(\cdot)$.

The values we visualised is computed by averaging AttentiveCAT scores from time steps that shared the same temporal position and feature bucket, noted as $V_{k,r,b}^{c}=\mathrm{mean}\left(s_t^c\mid\rho(t)=k,\operatorname{bucket}_r(\phi_r(x_t))=b\right)$. Here, $k$ denotes the temporal position, $r$ denotes the selected domain feature, and $b$ denotes the feature bucket. A positive $V_{k,r,b}^{c}$ indicates that the corresponding temporal-feature pattern provides evidence supporting class $c$, whereas a negative value indicates evidence against class $c$.

This formulation allows the visualisation to be adapted to different wearable health application by changing the temporal mapping, selected domain features, feature buckets setting, and target heath indicator. In the physical activity data example, time steps are mapped to hour of day, the selected domain features are MVPA, LPA and SED, the buckets setting as 10 minutes duration within an hour, health indicator as HWS. It is flexible to apply this formulation other wearable context.

\subsection{Experimental Evaluation}
We evaluated HealthCAT against three deep learning baselines (vanilla Transformer, GRU with attention, LSTM with attention) on both datasets. Each dataset was split 80\%/20\% at participant level. All models were trained using the Rectified Adam (RAdam) optimiser \cite{liu2019variance} for up to 50 epochs with early stopping. Hyperparameters were optimised for each model via 5-fold cross-validation grid search (Appendix~\ref{app:hyper_table}). Predictive performance was reported as F1-score (primary) and accuracy across 10 trials, with significance assessed by paired-sample $t$-test at $p<0.05$. 

To quantitatively evaluate interpretability (RQ2), we tested two hypotheses:
\begin{hyp} \label{hyp:first}
AttentiveCAT effectively selects the important time steps for prediction and consistently outperforms random selection.
\end{hyp}
\begin{hyp} \label{hyp:second}
HealthCAT consistently outperforms baseline methods across different data masking conditions.
\end{hyp}
Time steps within each participant's sequence were ranked by AttentiveCAT score; we then created masked subsets by zeroing out the features corresponding to the bottom 25\%, 50\%, and 75\%. Equivalent random-masking subsets served as controls. Each masking condition was evaluated over 100 trials, and significance was assessed by a paired-sample $t$-test at $p<0.05$.

\section{RESULTS}

\subsection{Predictive Performance (RQ1)}
In RQ1, we asked whether HealthCAT improves predictive performance over deep learning baselines for health-outcome prediction from wearable sensor data. Table~\ref{baseline_performance} reports the $F1$-score on both datasets across 10 trials, with hyperparameters selected via 5-fold grid-search cross-validation performed separately for each dataset (Appendix~\ref{app:hyper_table}). Accuracy was reported in \ref{app:pred_acc_table}. On the Physical Activity dataset, HealthCAT achieved a $F1$-score of $0.822 \pm 0.034$ and an accuracy of $0.826 \pm 0.027$, significantly outperforming all baseline methods (paired-sample $t$-test, $p<0.05$). On the DREAMT dataset, HealthCAT achieved an F1-score of $0.808 \pm 0.036$ and accuracy of $0.843 \pm 0.030$, again significantly exceeding all baselines. Performance was consistent across runs on both datasets, with low variability across trials.

\begin{table*}[!t]
  \caption{Comparison of Predictive Performance using F1-score and Accuracy(Higher is better).}
  
  \centering
  \begin{tabular}{cccccc}
    \toprule
    Dataset & Metrics & {HealthCAT (our approach)} & Transformer & GRU & LSTM \\
    \midrule
    Physical Activity & F1-score & \textbf{{$0.822\pm0.034$}\sig{*}} & $0.759 \pm 0.037$ & $0.742 \pm 0.051$ & $0.702 \pm 0.025$ \\
    {} & Accuracy & \textbf{{$0.826\pm0.027$}\sig{*}} & $0.784 \pm 0.026$ & $0.774 \pm 0.034$ & $0.738 \pm 0.018$ \\
    DREAMT & F1-score & \textbf{{$0.808\pm0.036$}\sig{*}} & $0.734 \pm 0.030$ & $0.749 \pm 0.052$ & $0.716 \pm 0.033$ \\
    {} & Accuracy & \textbf{{$0.843\pm0.030$}\sig{*}} & $0.788 \pm 0.042$ & $0.795 \pm 0.050$ & $0.780 \pm 0.035$ \\
    \bottomrule
  \end{tabular}
  \vspace{0.2em}
  {\par \raggedright Bold: Best Performance; *: Statistically significant improvement over  all baselines ($p < 0.05$) across 10 experiments. }
\label{baseline_performance}
\end{table*}

\subsection{Interpretability Evaluation (RQ2)}
In RQ2, we asked whether HealthCAT produces effective time-step-level interpretations, in which the identified time steps carry significantly more predictive information than do random selections. Table~\ref{table: ts_evaluation} reported $F1$-score under AttentiveCAT-guided masking and random masking; the corresponding accuracy table is provided in Appendix~\ref{app:pred_acc_table}. 

\begin{table*}[ht]
  \centering
  \caption{F1-score comparison between AttentiveCAT-guided masking and random masking across all models}
  \begin{tabular}{cccccc}
        \toprule
        & &
        \multicolumn{2}{c}{\bfseries Physical Activity dataset}&  
        \multicolumn{2}{c}{\bfseries DREAMT dataset} \\
        
        \cmidrule(lr){3-4}\cmidrule(lr){5-6}   
        {\makecell[c]{Methods}} & {Masking} & {AttentiveCAT} & {Random} &{AttentiveCAT} & {Random} \\

         & {Ratio} & {Selection} & {Selection} & {Selection} & {Selection} \\
     
        \midrule
        HealthCAT & 
        25\% & \textbf{{$0.7277\pm0.0048$}\sig{*\,†§}} & $0.6947\pm0.0053$ & \textbf{{$0.7204\pm0.0073$}\sig{*\,†§}}& $0.6993\pm0.0114$ \\
        
        (our approach) & 50\% & \textbf{{$0.7183\pm0.0059$}\sig{*\,†§}} & $0.6685\pm0.0055$ & \textbf{{$0.7134\pm0.0118$}\sig{*\,†§}} & $0.6743\pm0.0080$ \\
    
        & 75\% & \textbf{{$0.7578\pm0.0043$}\sig{*\,†§}}&  $0.6349\pm0.0055$ & \textbf{{$0.7252\pm0.0096$}\sig{*\,†§}} & $0.6590\pm0.0078$\\
    
        Transformer  & 
        25\% & $0.7231\pm0.0050$*  & $0.6730\pm0.0041$ & $0.7221\pm0.0094$& $0.6980\pm0.0093$\\
    
        & 50\% & $0.7122\pm0.0062$* & $0.6620\pm0.0076$ & $0.6930\pm0.0100$& $0.6808\pm0.0066$\\
    
        & 75\% & $0.7480\pm0.0052$* & $0.6369\pm0.0040$ & $0.7158\pm0.0081$* & $0.6436\pm0.0024$\\
    
        GRU &
        25\% & $0.6793\pm0.0046$* & $0.6533\pm0.0070$ & $0.6369\pm0.0103$ & $0.6203\pm0.0121$\\
    
        & 50\% & $0.5912\pm0.0014$ & $0.5942\pm0.0014$ & $0.6020\pm0.0096$& $0.5883\pm0.0084$\\
    
        & 75\% & $0.7036\pm0.0048$* & $0.5938\pm0.0010$ & $0.6301\pm0.0127$* & $0.5585\pm0.0080$\\
    
        LSTM &
        25\% & $0.6619\pm0.0069$* & $0.6533\pm0.0070$ &$0.6409\pm0.0112$ & $0.6204\pm0.0087$\\
    
        & 50\% & $0.5914\pm0.0012$ & $0.5941\pm0.0012$ & $0.6355\pm0.0124$& $0.6160\pm0.0121$\\
    
        & 75\% & $0.6511\pm0.0064$* & $0.5933\pm0.0009$ &$0.6407\pm0.0091$* & $0.5600\pm0.0097$\\
        
        \bottomrule

      \end{tabular}
      \vspace{0.2em}

      {\par \raggedright Results report mean ± standard error over 100 experiments. }
      {\par \raggedright *: Significant improvement of AttentiveCAT selection over Random selection ($p < 0.05$, paired-sample $t$-test). }
      {\par \raggedright \sig{\,†§}: Significant improvement of HealthCAT (our approach) over GRU and LSTM, respectively ($p < 0.05$, paired-sample $t$-test). }
    \label{table: ts_evaluation}
\end{table*}

In the Physical Activity dataset, AttentiveCAT-guided masking obtained significantly higher performance than random masking in 10 out of 12 model-masking comparisons ($p < 0.05$), with the largest differences observed at 75\% masking level. At the most aggressive masking setting, models predicted using only the top 25\% of AttentiveCAT selected time steps, solely relying on the most informative temporal segments. At a 50\% masking level, AttentiveCAT consistently produced positive mean differences compared with random selections. When masking at 25\%, AttentiveCAT selection still significantly outperformed random selection across all models ($p < 0.05$). On the other hand, in the DREAMT dataset, significance was observed at a 75\% masking level. AttentiveCAT selection consistently achieved a higher F1-score than random selection in multiple models, including GRU: $t(100)$ = 2.016, $p$=0.023 and Transformer $t(100)$ = 1.73, $p$=0.043. It has been shown that the time steps ranked highest by AttentiveCAT selection retain predictive information even when a large proportion of the sequence is masked. At 50\% and 25\% masking levels, there were no instances where random selection outperformed AttentiveCAT selection, although fewer significant effects were observed. Combined with the paired-sample $t$-test results from the PA dataset, there is evidence supporting our hypothesis $H1$ that AttentiveCAT effectively selects important time steps and consistently outperforms random selection across datasets. 

\subsection{Domain-Adapted Visualisation (RQ3)}

In RQ3, we asked how can HealthCAT's time-step-level importance score facilitate to surface behavioural patterns that is meaningful to the given domain. Following the description in Section~\ref{sec:method_rq3}, we visualised the contrastive participants' interpretations across MVPA, LPA, and SED on a 24-hour cycle (Fig.~\ref{fig:case_study_2in1}) as illustrative examples. The domain-adapted visualisations show when and at what intensity level the temporal patterns contribute to the predicted category.

\begin{itemize}
  \item \textbf{Participant A} is classified as being in the Healthy Fitness Zone category, as indicated in both the ground truth and the predicted label; 
  \item \textbf{Participant B} is classified as being in the Need Improvement category, as indicated in both the ground truth and the predicted label; 
\end{itemize}

For each participant, time-step-level importance scores were mapped back onto the original hourly PA time series and visualised with respect to MVPA, LPA, and SED, as shown in Fig.~\ref{fig:case_study_2in1}. These scores are computed with respect to the participant’s predicted class; thus, positive values indicate evidence supporting that predicted class (Healthy Fitness Zone for Participant A, Need Improvement for Participant B), while negative values indicate counter evidence. Each subfigure in Fig.~\ref{fig:case_study_2in1} represents the importance score associated with the time spent at a specific PA intensity level (y-axis) across time of the day (x-axis). This visualisation enables users to inspect when specific activity intensities contribute most strongly to the health indicator prediction which is HWS this case study.

\begin{figure*}[t]
    \centering
    \textbf{Participant A — HealthCAT Domain-Adapted Visualisation}
    \vspace{0.4em}

    \begin{subfigure}{\linewidth}
    \caption{\textbf{MVPA} Time-of-Day Contributions to the Healthy Fitness Zone Category Prediction}
    \includegraphics[width=\linewidth,height=0.28\textheight,keepaspectratio]{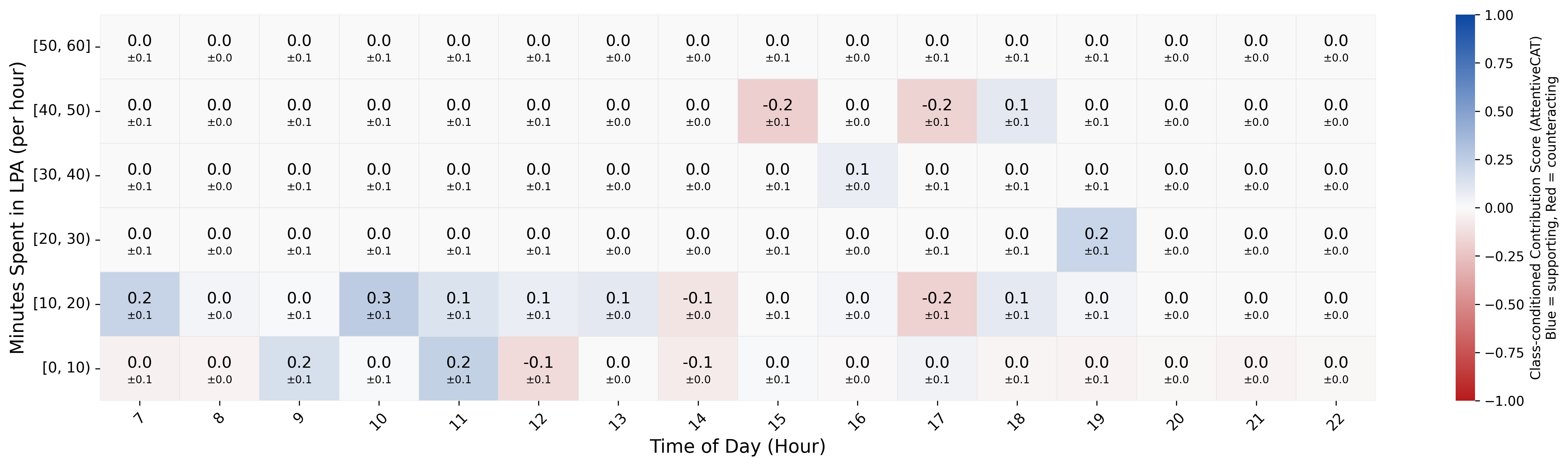}
    \end{subfigure}
    
    \begin{subfigure}{\linewidth}
    \caption{\textbf{LPA} Time-of-Day Contributions to the Healthy Fitness Zone Category Prediction}
    \includegraphics[width=\linewidth,height=0.28\textheight,keepaspectratio]{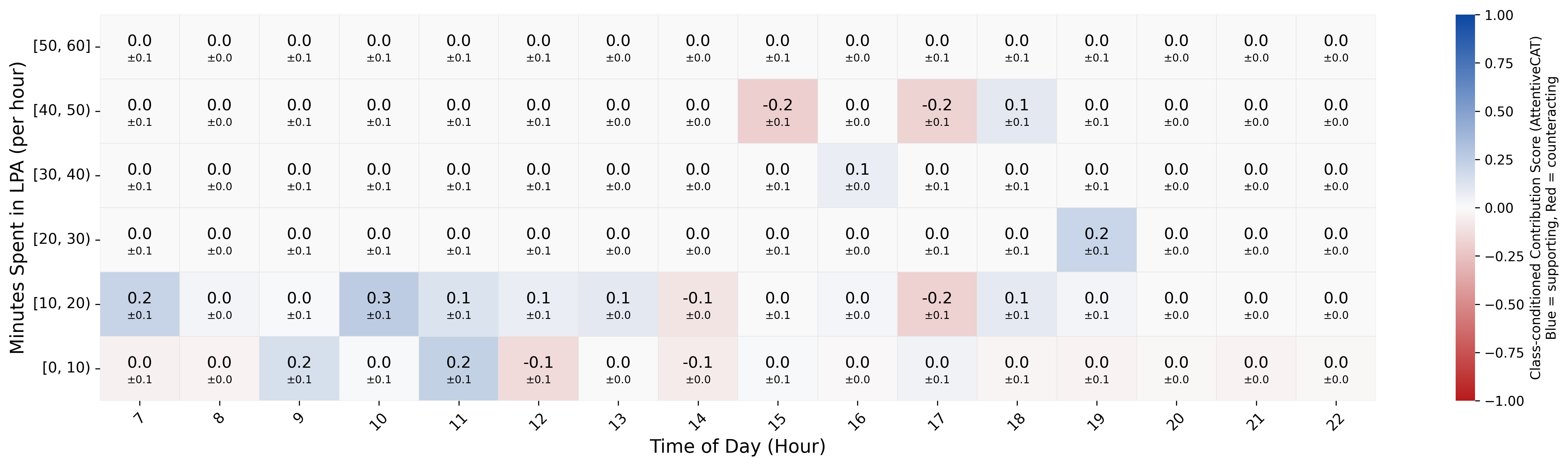}
    \end{subfigure}
    
    \begin{subfigure}{\linewidth}
    \caption{\textbf{SED} Time-of-Day Contributions to the Healthy Fitness Zone Category Prediction}
    \includegraphics[width=\linewidth,height=0.28\textheight,keepaspectratio]{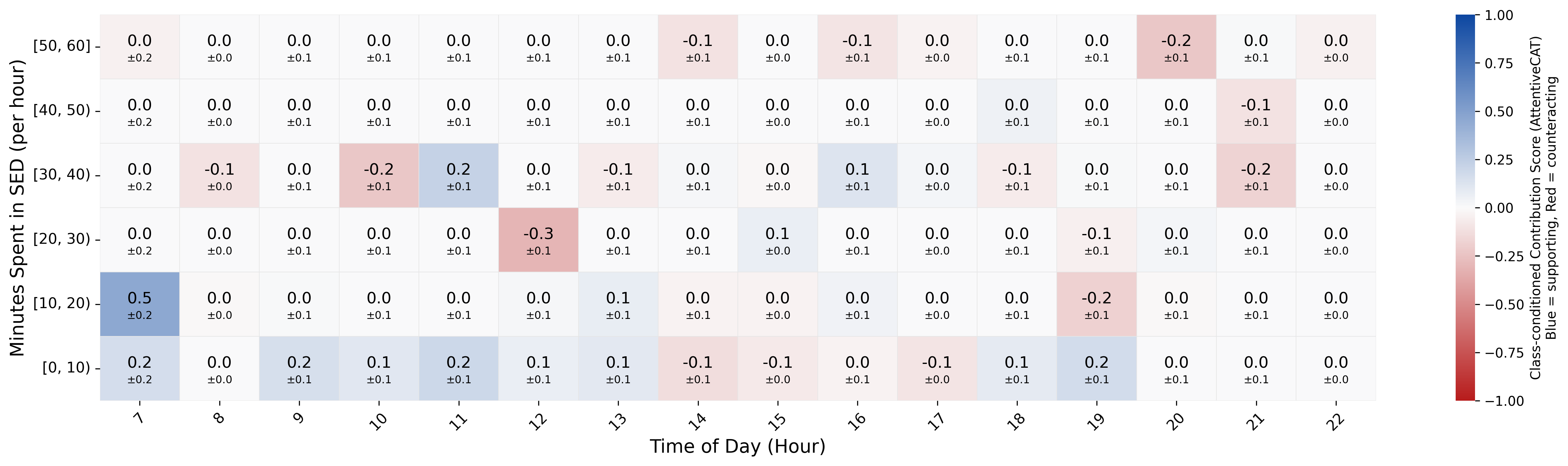}
    \end{subfigure}
    
    \caption{(a)(b)(c). Time-of-day visualisations of class-conditioned contributions produced by HealthCAT for \textbf{Participant A}, whose data were predicted as belonging to the Healthy Fitness Zone category. Each sub-visualisation shows domain-adapted AttentiveCAT importance scores computed with respect to the predicted class. Positive values (blue) indicate time segments that provide evidence supporting the Healthy Fitness Zone prediction, while negative values (red) indicate segments that counteract the prediction. Sub-visualisations are shown for MVPA, LPA, and SED (sleeping time excluded), with activity duration binned by minutes per hour.}
\end{figure*}

\begin{figure*}[t]
    \ContinuedFloat
    \centering
    \textbf{Participant B — HealthCAT Domain-Adapted Visualisation}

    \vspace{0.4em}

    \begin{subfigure}{\linewidth}
    \caption{\textbf{MVPA} Time-of-Day Contributions to the Need Improvement Category Prediction}
    \includegraphics[width=\linewidth,height=0.28\textheight,keepaspectratio]{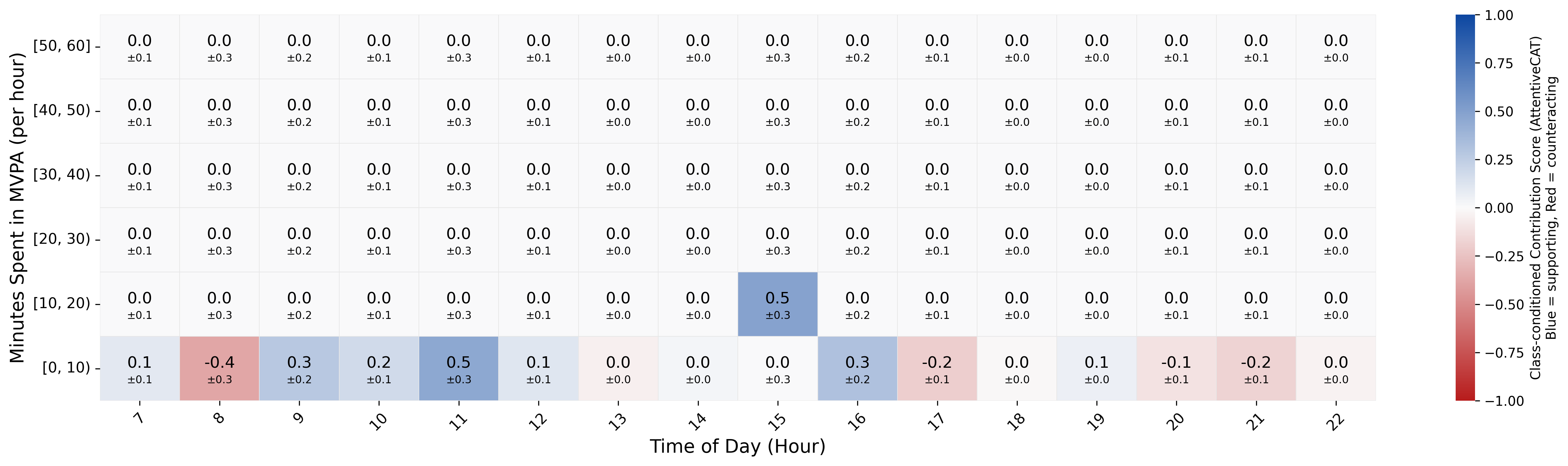}
    \end{subfigure}
    
    \begin{subfigure}{\linewidth}
    \caption{\textbf{LPA} Time-of-Day Contributions to the Need Improvement Category Prediction}
    \includegraphics[width=\linewidth,height=0.28\textheight,keepaspectratio]{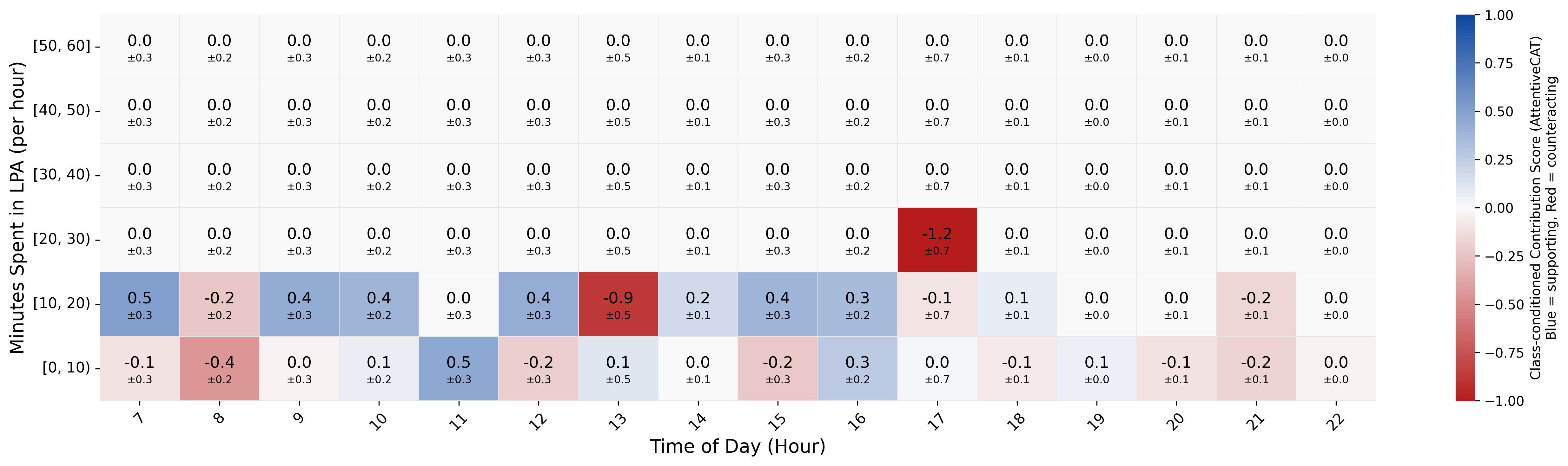}
    \end{subfigure}
    
    \begin{subfigure}{\linewidth}
    \caption{\textbf{SED} Time-of-Day Contributions to the Need Improvement Category Prediction}
    \includegraphics[width=\linewidth,height=0.28\textheight,keepaspectratio]{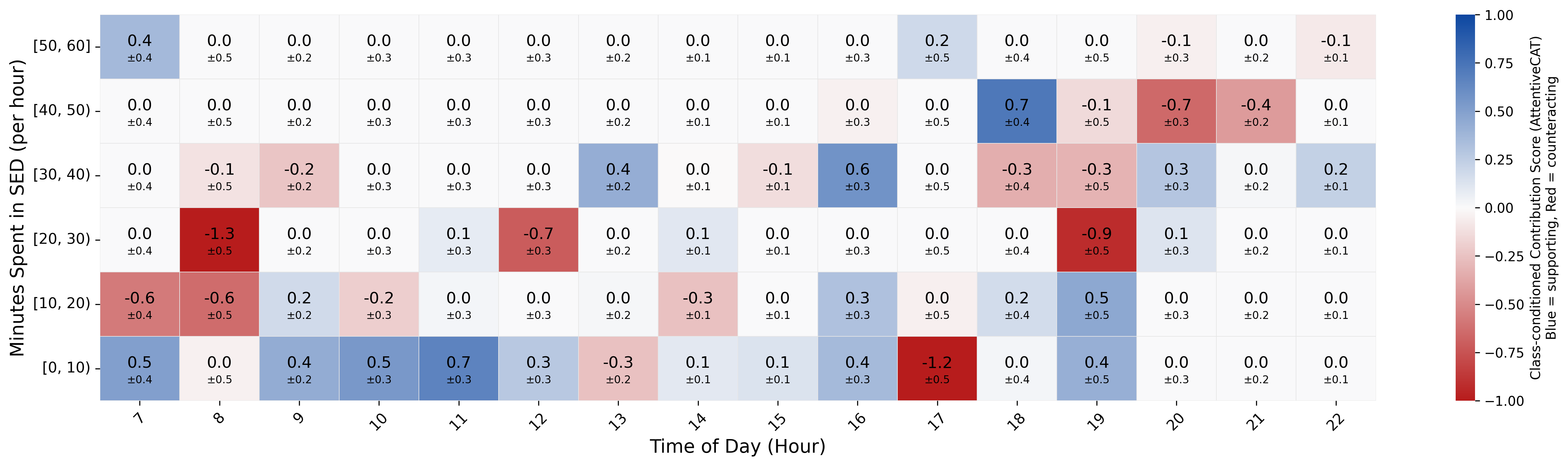}
    \end{subfigure}

    \caption{(d)(e)(f). Time-of-day visualisations of class-conditioned contributions produced by HealthCAT for \textbf{Participant B}, whose data were predicted as belonging to the Need Improvement category. Each sub-visualisation shows domain-adapted AttentiveCAT importance scores computed with respect to the predicted class. Positive values (blue) indicate time segments that provide evidence supporting the Need Improvement prediction, while negative values (red) indicate segments that counteract the prediction. Sub-visualisations are shown for MVPA, LPA, and SED (sleeping time excluded), with activity duration bucketed by minutes per hour.}
    \label{fig:case_study_2in1}
\end{figure*}

Distinct individual-level patterns were observed across the two participants. For Participant A, a high positive importance score appeared at 9:00 (subfigure (a), 40 to 50 minutes), continuous positive importance score for LPA from 10:00 to 13:00 (subfigure (b)), and negative importance scores for SED at several time points: 10:00, 12:00, 14:00, 17:00, 20:00, and 21:00 (subfigure (c)). For Participant B, MVPA  was limited to short durations between 9:00 and 11:00 (less than 10 minutes) and only a burst at 15:00 (subfigure (d)); LPA consistently remained below 20 minutes across most hours (subfigure (e)). Positive importance scores for SED appeared at 7:00, 11:00, 13:00, 16:00, and 18:00 (subfigure (f)).

\section{DISCUSSION}
 
This study addressed a core challenge in wearable health analytics: how to combine the predictive strength of deep learning models with time-aware interpretability for individualised health analysis (e.g. everyday behavioural patterns). Our findings suggest that HealthCAT can support personalised and predictive healthcare by identifying when specific physiological or behavioural patterns contribute to individual-level health predictions. The contribution of HealthCAT is not limited to improving classification performance. In many wearable health applications, aggregated features such as daily average MVPA, daily total steps, or average sleeping hours are useful but not comprehensive enough to understand the underlying behavioural patterns. HealthCAT moves beyond these aggregate representations by identifying temporal evidence associated with health outcomes. This is particularly relevant to intelligent informatics for personalised healthcare, where models are expected not only to predict outcomes but also to provide evidence that can be inspected, validated, and potentially translated into health monitoring or intervention design. The following sections provide discussions of each research question in the study. 

\subsection{Predictive Performance (RQ1)} 

The results of RQ1 demonstrate the predictive power of HealthCAT for wearable health modelling. HealthCAT consistently outperformed the baselines (Transformer, LSTM, and GRU) on both the free-living Physical Activity dataset and the laboratory-recorded DREAMT dataset, with improvements of up to 17\% in F1-score and up to 12\% in accuracy. The improvement in prediction is also driven by the sensor modality aware preprocessing and Encoder-only structure \ref{fig:overview_method}. The preprocessing step transformed the raw sensor streams into structured multivariate time series at meaningful temporal resolutions for each dataset, such as hourly physical activity features and minute-level sleep features. 

In terms of prediction, our study provides evidence that structured wearable sensor data can support health outcome prediction. HealthCAT offers opportunities to shift the focus of deep learning applications from recognising activity states, sleep stages, or physiological events \cite{supratak2017deepsleepnet, rajpurkar2017cardiologist, sharma2022transformer, singh2020deep, yu2023semi} to predicting broader individual-level health indicators. Taken together, the contribution of HealthCAT is not simply to apply a deep learning method to wearable sensor data; it provides a flexible modelling framework in which the sensor modality, temporal granularity, and target health outcome can be adjusted according to the available data and the intended health application. 

\subsection{Interpretability Evaluation (RQ2)}

The results of RQ2 show that HealthCAT's time-step-level importance scores are not only interpretable but also predictively informative. Our proposed interpretability evaluation protocol provides evidence that HealthCAT's explanations are closely tied to the model's decisions regarding predictions. HealthCAT outperformed all baseline models in the cumulative target masking experiments related to RQ2, thereby supporting Hypothesis~\ref{hyp:second}. The results confirm that HealthCAT efficiently identifies time steps impacting the model's decisions and provides significantly more predictive information than randomly selected time steps, even under aggressive masking levels. This result supports Hypothesis~\ref{hyp:first}. 

HealthCAT addresses an important interpretability limitation in current wearable health modelling. Many deep learning applications report strong predictive performance, but their explanations are either absent, qualitative, or difficult to evaluate \cite{singh2020deep,verde2022wearable, yu2023semi}. HealthCAT leverages attention weights together with class activation gradients to expose the time-step-level contribution to the prediction. By proposing a targeted masking evaluation method to compare HealthCAT selected important time steps with random selections, we quantitatively evaluated whether the identified time steps carry predictive information. 

In the experiments, a non-monotonic pattern was observed under HealthCAT-guided masking, where performance declines as low-importance time steps are removed at 25\% and 50\%, but recovers at 75\% masking. We interpret this recovery as a noise-filtering robustness within the framework; with aggressive masking, the model is forced to rely on the most informative time steps selected by HealthCAT, while less informative segments are filtered out. By contrast, random masking degrades monotonically, as it indiscriminately removes both informative and uninformative time steps. 

Together, these results indicate that HealthCAT's interpretable extraction block (in Fig.~\ref{fig:overview_method}) identifies temporal patterns that genuinely contribute to the model's predictive decisions, rather than reflecting arbitrary attention allocation. By spotlighting class-specific, time-step-level importance, we address the limitations of existing interpretability approaches for wearable data, which often lack temporal specificity or fail to reflect the model's decisions. Furthermore, HealthCAT contributes a new form of temporal interpretation for wearable health analytics, which may help health researchers focus on the most informative periods within long wearable recordings. This is particularly important in health-related research settings, where interpretations need to be inspected and evaluated before they can support decision-making. 

\subsection{Domain-Adapted Visualisation (RQ3)}\label{result_qual}

The illustrative visualisations in RQ3 \ref{sec:method_rq3} demonstrate how time-step-level importance score facilitate domain-adapted visualisation that surface behavioural patterns meaningful to the target domain. Existing deep learning applications on wearable data have introduced attention-based methods to improve model's interpretability \cite{phan2022sleeptransformer, jin2024cardiologist, garcia2024sleepecg, abdelaal2024exploring}. However, these interpretations are not always translated into domain-relevant behavioural patterns. HealthCAT extends prior work by connecting model-derived importance scores with time-of-day and domain feature dimensions, allowing stakeholders (e.g., health researchers, clinicians, and individuals) to inspect individual behavioural patterns in a domain-relevant form.

In the physical activity data examples, the identified importance scores were projected back to a 24-hour cycle (excluding sleeping time) by activity intensities (MVPA, LPA) and SED. The visualisations make the model output easier to inspect because they align the interpretations with when physical activity occurs and at which intensity level, and they also reflect the contributions to the health outcome.

The two contrastive participants show the value of this domain adaptation. From the holistic overview of Fig.~\ref{fig:case_study_2in1}, we observed distinct patterns in the two participants' visualisations across MVPA, LPA and SED. Participant~A showed a more structured activity profile, where sustained morning MVPA and consistent late morning to early afternoon LPA provided supportive evidence for the Healthy Fitness Zone category. In contrast, Participant~B showed a more fragmented activity profile, with limited MVPA, low LPA across most hours, and several sedentary periods contributing to the Need Improvement category prediction. Importantly, sedentary behavioural was not uniformly interpreted as "bad", where adjacent sedentary periods sometimes showed opposite contributions. This observation tackled an important consideration that sedentary time is context-dependent, and explanations need to be understood in relation to routine and setting rather than being treated as universally "bad". This pattern is consistent with prior accelerometer-based study showing the importace of sedentary accumulation and temporal context in health outcomes \cite{ekelund2019dose}. 

The domain-adapted visualisations of HealthCAT are not hard-coded for a fixed physical activity specific output. The projection approach introduced in Section~\ref{sec:method_rq3} can be applied to other wearable contexts (e.g., minute-level sleep features) by selecting  appropriate variables from the data. HealthCAT can therefore support behavioural pattern inspection, hypothesis generation, and health intervention design. 

\subsection{Limitation and Future Work}
Several limitations should be acknowledged. First, although the two datasets differ in temporal structure, the evaluation remains limited in terms of sample size and cohort diversity. Second, this study focuses solely on wearable sensor data; however, contextual variables (e.g., dietary intake, smoking, and demographics) are also key factors in health outcomes \cite{uddin2020clustering}. Future work will extend HealthCAT in three directions: (i) evaluation on larger and more diverse cohorts; (ii) integration of multimodal contextual metadata (e.g., dietary intake) for more comprehensive health analytics; and (iii) an interactive application that can be practically used in health interventions, drawing on previous work such as HealthPrism \cite{jiang2023healthprism}.

\section{CONCLUSION}

We presented HealthCAT, an interpretable Encoder-only Transformer framework that produces time-step-level, class-specific interpretations for health predictions from wearable sensor data. Evaluated across two real-world datasets, HealthCAT outperformed the baselines in both predictive and interpretability validation, with masking experiments confirming that the identified time steps offer significantly greater predictive value than random selection. By mapping importance scores onto domain-relevant behavioural cycles, HealthCAT moves beyond aggregated summary metrics to reveal when and how physical activity patterns occur at the individual level. This positions HealthCAT as a practical tool for health researchers seeking richer, temporally grounded insights to support personalised health intervention design, long-term health monitoring, and personalised behaviour recommendations.

\section*{References}

\clearpage
\appendices

\section{Hyperparameters}
\label{app:hyper_table}
\begin{table}[ht]
  
  \centering
  \begin{tabular}{ll}
    \toprule
 
    Method & Hyperparameters \\

    \midrule
    HealthCAT & learning rate: 0.001, 0.0005, \textbf{\underline{0.0001}} \\
              & number of layers: \textbf{\underline{3}}, 6, 8 \\
              & number of heads: 2, \textbf{\underline{4}}, 8 \\
              & model dimension: 16, \textbf{32}, \underline{64}, 128 \\
              & batch size: 8, \textbf{\underline{16}}, 32 \\   
    \midrule
    Transformer & learning rate: 0.001, 0.0005, \textbf{\underline{0.0001}} \\
                & number of layers: \textbf{\underline{3}}, 6, 8 \\
                & number of heads: 2, \textbf{\underline{4}}, 8 \\
                & model dimension: 16, \textbf{32}, \underline{64}, 128 \\
                & batch size: 8, \textbf{\underline{16}}, 32 \\   
    \midrule
    GRU & learning rate: 0.001, 0.0005, \textbf{\underline{0.0001}} \\
              & number of layers: 2, \textbf{\underline{3}}, 6 \\
              & model dimension: 16, \textbf{\underline{32}}, 64, 128 \\
              & batch size: 8, \textbf{\underline{16}}, 32 \\   
              & dropout: 0.1, \textbf{0.2}, \underline{0.3} \\

    \midrule
    LSTM & learning rate: 0.001, 0.0005, \textbf{\underline{0.0001}}  \\
              & number of layers: 2, \textbf{\underline{3}}, 6 \\
              & model dimension: 16, \underline{32}, \textbf{64}, 128 \\
              & batch size: \underline{8},\textbf{16}, 32  \\   
              & dropout: 0.1, \textbf{0.2}, \underline{0.3} \\

    \bottomrule
  \end{tabular}
  \vspace{0.2em}
  {\par \raggedright The best parameters (annotated) are obtained from grid search cross-validation with AdamW as optimiser.}
  {\par \raggedright Best parameters for the Physical Activity dataset are highlighted in \textbf{bold}, and for the DREAMT dataset are annotated in \underline{underline}.}
\end{table}

\section{Accuracy Comparison Between AttentiveCAT-guided Masking and Random Masking Across All Methods}
\label{app:pred_acc_table}
\begin{table}[ht]

  \centering
  \newcommand{\pa}[1]{\eqmakebox[CFNetvsNeuMF][r]{#1}}
  \newcommand{\hlc}{\textbf}
  \begin{tabular}{cccccc}
        \toprule
        & &
        \multicolumn{2}{c}{\bfseries Physical Activity dataset}&  
        \multicolumn{2}{c}{\bfseries DREAMT dataset} \\
        
        \cmidrule(lr){3-4}\cmidrule(lr){5-6}   
        {\makecell[c]{Methods}} & {Masking} & {AttentiveCAT} & {Random} &{AttentiveCAT} & {Random} \\

         & {Ratio} & {Selection} & {Selection} & {Selection} & {Selection} \\
     
        \midrule

        HealthCAT & 
        25\% & $ 0.7490 \pm 0.0044 $*  & $ 0.7287 \pm 0.0053 $ & $ 0.7680\pm0.0066 $& $0.7526 \pm0.0096 $\\
    
        (our approach) & 50\% & {$0.7384\pm0.0056$}\sig{*\,†} & $0.7185 \pm 0.0048 $ & $0. 7649\pm0.0102 $*& $0.7307\pm0.0074$\\
    
        & 75\% & $0.7758 \pm 0.0039$* & $0.6579 \pm 0.0048$ & $0.7678\pm 0.0082$* & $0.7253\pm0.0069$\\
    
        Transformer  & 
        25\% & $ 0.7490 \pm 0.0044 $*  & $ 0.7232 \pm 0.0038 $ & $ 0.7728\pm0.0076 $*& $0.7458 \pm0.0079 $\\
    
        & 50\% & $0.7494 \pm0.0057 $* & $0.7049 \pm0.0044 $ & $0.7475 \pm0.0083 $& $0.7405\pm0.0058$\\
    
        & 75\% & $0.7661\pm0.0045$* & $0.6584\pm0.0050$ & $0.7644\pm0.0069$* & $0.7090\pm0.0106$\\
    
        GRU &
        25\% & $0.7429\pm0.0028$* & $0.7087\pm0.0013$ & $0.7431\pm0.0079$ & $0.7288\pm0.0093$\\
    
        & 50\% & $0.7087\pm0.0007$* & $0.7025\pm0.0016$ & $0.7210\pm0.0073$* & $0.7055\pm0.0066$\\
    
        & 75\% & $0.7509\pm0.0038$* & $0.6365\pm0.0042$ & $0.7376\pm0.0095$* & $0.6850\pm0.0062$\\
    
        LSTM &
        25\% & $0.7281\pm0.0036$* & $0.7097\pm0.0000$ &$0.7465\pm0.0084$ & $0.7320\pm0.0066$\\
    
        & 50\% & $0.7081\pm0.0011$* & $0.6939\pm0.0034$ & $0.7421\pm0.0093$& $0.7275\pm0.0091$\\
    
        & 75\% & $0.7090\pm0.0031$* & $0.6552\pm0.0056$ &$0.7470\pm0.0069$* & $0.6855\pm0.0076$\\
        
        \bottomrule

      \end{tabular}
      \vspace{0.2em}

      {\par \raggedright Results report mean ± standard error over 100 experiments. }
      {\par \raggedright *: Significant improvement of AttentiveCAT selection over Random selection ($p$ < 0.05, paired-sample $t$-test). }
      {\par \raggedright \sig{\,†}: Significant improvement of HealthCAT (our approach) over GRU. ($p$ < 0.05, paired-sample $t$-test). }
\end{table}


\begin{thebibliography}{10}
\providecommand{\url}[1]{#1}
\csname url@samestyle\endcsname
\providecommand{\newblock}{\relax}
\providecommand{\bibinfo}[2]{#2}
\providecommand{\BIBentrySTDinterwordspacing}{\spaceskip=0pt\relax}
\providecommand{\BIBentryALTinterwordstretchfactor}{4}
\providecommand{\BIBentryALTinterwordspacing}{\spaceskip=\fontdimen2\font plus
\BIBentryALTinterwordstretchfactor\fontdimen3\font minus
  \fontdimen4\font\relax}
\providecommand{\BIBforeignlanguage}[2]{{%
\expandafter\ifx\csname l@#1\endcsname\relax
\typeout{** WARNING: IEEEtran.bst: No hyphenation pattern has been}%
\typeout{** loaded for the language `#1'. Using the pattern for}%
\typeout{** the default language instead.}%
\else
\language=\csname l@#1\endcsname
\fi
#2}}
\providecommand{\BIBdecl}{\relax}
\BIBdecl

\bibitem{ravi2016deep}
D.~Ravi, C.~Wong, B.~Lo, and G.-Z. Yang, ``A deep learning approach to on-node
  sensor data analytics for mobile or wearable devices,'' \emph{IEEE journal of
  biomedical and health informatics}, vol.~21, no.~1, pp. 56--64, 2016.

\bibitem{khurshid2022wearable}
S.~Khurshid, L.-C. Weng, V.~Nauffal, J.~P. Pirruccello, R.~A. Venn, M.~A.
  Al-Alusi, E.~J. Benjamin, P.~T. Ellinor, and S.~A. Lubitz, ``Wearable
  accelerometer-derived physical activity and incident disease,'' \emph{NPJ
  Digital Medicine}, vol.~5, no.~1, p. 131, 2022.

\bibitem{henriksen2018using}
A.~Henriksen, M.~Haugen~Mikalsen, A.~Z. Woldaregay, M.~Muzny, G.~Hartvigsen,
  L.~A. Hopstock, and S.~Grimsgaard, ``Using fitness trackers and smartwatches
  to measure physical activity in research: analysis of consumer wrist-worn
  wearables,'' \emph{Journal of medical Internet research}, vol.~20, no.~3, p.
  e110, 2018.

\bibitem{braem2025association}
C.~Braem, W.~Pasman, T.~van~den Broek, M.~Caspers, F.~Jagers, U.~Yavuz,
  H.~Hermens, P.~Veltink, and S.~Wopereis, ``The association of physical
  activity, heart rate and sleep from an activity tracker with weight loss
  during a 6-month personalized combined lifestyle intervention: a
  retrospective analysis,'' \emph{BMC Digital Health}, vol.~3, no.~1, p.~8,
  2025.

\bibitem{moshawrab2023smart}
M.~Moshawrab, M.~Adda, A.~Bouzouane, H.~Ibrahim, and A.~Raad, ``Smart wearables
  for the detection of cardiovascular diseases: a systematic literature
  review,'' \emph{Sensors}, vol.~23, no.~2, p. 828, 2023.

\bibitem{ahmed2022overview}
A.~Ahmed, S.~Aziz, A.~Abd-Alrazaq, F.~Farooq, and J.~Sheikh, ``Overview of
  artificial intelligence--driven wearable devices for diabetes: scoping
  review,'' \emph{Journal of medical Internet research}, vol.~24, no.~8, p.
  e36010, 2022.

\bibitem{lubitz2022detection}
S.~A. Lubitz, A.~Z. Faranesh, C.~Selvaggi, S.~J. Atlas, D.~D. McManus, D.~E.
  Singer, S.~Pagoto, M.~V. McConnell, A.~Pantelopoulos, and A.~S. Foulkes,
  ``Detection of atrial fibrillation in a large population using wearable
  devices: the fitbit heart study,'' \emph{Circulation}, vol. 146, no.~19, pp.
  1415--1424, 2022.

\bibitem{ferguson2022effectiveness}
T.~Ferguson, T.~Olds, R.~Curtis, H.~Blake, A.~J. Crozier, K.~Dankiw, D.~Dumuid,
  D.~Kasai, E.~O'Connor, R.~Virgara \emph{et~al.}, ``Effectiveness of wearable
  activity trackers to increase physical activity and improve health: a
  systematic review of systematic reviews and meta-analyses,'' \emph{The Lancet
  Digital Health}, vol.~4, no.~8, pp. e615--e626, 2022.

\bibitem{jorke2025gptcoach}
M.~J{\"o}rke, S.~Sapkota, L.~Warkenthien, N.~Vainio, P.~Schmiedmayer,
  E.~Brunskill, and J.~A. Landay, ``Gptcoach: Towards llm-based physical
  activity coaching,'' in \emph{Proceedings of the 2025 CHI Conference on Human
  Factors in Computing Systems}, 2025, pp. 1--46.

\bibitem{WHO_BeActive}
\BIBentryALTinterwordspacing
{World Health Organization}, ``Be active: Physical activity,'' 2025, accessed:
  2025-08-12. [Online]. Available:
  \url{https://www.who.int/initiatives/behealthy/physical-activity}
\BIBentrySTDinterwordspacing

\bibitem{phan2021xsleepnet}
H.~Phan, O.~Y. Ch{\'e}n, M.~C. Tran, P.~Koch, A.~Mertins, and M.~De~Vos,
  ``Xsleepnet: Multi-view sequential model for automatic sleep staging,''
  \emph{IEEE Transactions on Pattern Analysis and Machine Intelligence},
  vol.~44, no.~9, pp. 5903--5915, 2021.

\bibitem{perslev2021u}
M.~Perslev, S.~Darkner, L.~Kempfner, M.~Nikolic, P.~J. Jennum, and C.~Igel,
  ``U-sleep: resilient high-frequency sleep staging,'' \emph{NPJ digital
  medicine}, vol.~4, no.~1, p.~72, 2021.

\bibitem{raza5021631lightweight}
A.~Raza, K.~P. Tran, L.~Koehl, X.~Zeng, K.~Benzaidi, S.~Li, S.~Hotham, and
  K.~D. Tran, ``Lightweight transformer for federated human activity
  recognition in home care,'' \emph{Available at SSRN 5021631}, 2021.

\bibitem{khan2022capslstm}
P.~Khan, Y.~Kumar, and S.~Kumar, ``Capslstm-based human activity recognition
  for smart healthcare with scarce labeled data,'' \emph{IEEE Transactions on
  Computational Social Systems}, vol.~11, no.~1, pp. 707--716, 2022.

\bibitem{10064701}
T.~Ahmad and J.~Wu, ``Sdigru: Spatial and deep features integration using
  multilayer gated recurrent unit for human activity recognition,'' \emph{IEEE
  Transactions on Computational Social Systems}, vol.~11, no.~1, pp. 973--985,
  2024.

\bibitem{verde2022wearable}
L.~Verde and G.~Sannino, ``Wearable sensor signals: An overview of the ai
  models most commonly applied to time series data analysis,'' \emph{Connected
  e-Health: Integrated IoT and Cloud Computing}, pp. 147--163, 2022.

\bibitem{miao2024spatial}
S.~Miao, L.~Chen, and R.~Hu, ``Spatial-temporal masked autoencoder for
  multi-device wearable human activity recognition,'' \emph{Proceedings of the
  ACM on Interactive, Mobile, Wearable and Ubiquitous Technologies}, vol.~7,
  no.~4, pp. 1--25, 2024.

\bibitem{zhou2024autoaughar}
Y.~Zhou, H.~Zhao, Y.~Huang, T.~R{\"o}ddiger, M.~Kurnaz, T.~Riedel, and
  M.~Beigl, ``Autoaughar: automated data augmentation for sensor-based human
  activity recognition,'' \emph{Proceedings of the ACM on Interactive, Mobile,
  Wearable and Ubiquitous Technologies}, vol.~8, no.~2, pp. 1--27, 2024.

\bibitem{zerveas2021transformer}
G.~Zerveas, S.~Jayaraman, D.~Patel, A.~Bhamidipaty, and C.~Eickhoff, ``A
  transformer-based framework for multivariate time series representation
  learning,'' in \emph{Proceedings of the 27th ACM SIGKDD conference on
  knowledge discovery \& data mining}, 2021, pp. 2114--2124.

\bibitem{qiang2022attcat}
Y.~Qiang, D.~Pan, C.~Li, X.~Li, R.~Jang, and D.~Zhu, ``Attcat: Explaining
  transformers via attentive class activation tokens,'' \emph{Advances in
  neural information processing systems}, vol.~35, pp. 5052--5064, 2022.

\bibitem{wattelez2025assessing}
G.~Wattelez, S.~Frayon, and O.~Galy, ``Assessing physical activity/behavior of
  adolescents living in the pacific with accelerometer data: 231 geneactiv
  records in new caledonia,'' \emph{Data in Brief}, vol.~58, p. 111228, 2025.

\bibitem{wang2024dreamt}
K.~Wang, J.~Yang, A.~Shetty, and J.~Dunn, ``Dreamt: Dataset for real-time sleep
  stage estimation using multisensor wearable technology,'' \emph{PhysioNet
  https://doi. org/10.13026/62AN-CB28}, 2024.

\bibitem{singh2020deep}
S.~P. Singh, M.~K. Sharma, A.~Lay-Ekuakille, D.~Gangwar, and S.~Gupta, ``Deep
  convlstm with self-attention for human activity decoding using wearable
  sensors,'' \emph{IEEE Sensors Journal}, vol.~21, no.~6, pp. 8575--8582, 2020.

\bibitem{liu2021gated}
M.~Liu, S.~Ren, S.~Ma, J.~Jiao, Y.~Chen, Z.~Wang, and W.~Song, ``Gated
  transformer networks for multivariate time series classification,''
  \emph{arXiv preprint arXiv:2103.14438}, 2021.

\bibitem{wen2022transformers}
Q.~Wen, T.~Zhou, C.~Zhang, W.~Chen, Z.~Ma, J.~Yan, and L.~Sun, ``Transformers
  in time series: A survey,'' \emph{arXiv preprint arXiv:2202.07125}, 2022.

\bibitem{le2024shapeformer}
X.-M. Le, L.~Luo, U.~Aickelin, and M.-T. Tran, ``Shapeformer: Shapelet
  transformer for multivariate time series classification,'' in
  \emph{Proceedings of the 30th ACM SIGKDD Conference on Knowledge Discovery
  and Data Mining}, 2024, pp. 1484--1494.

\bibitem{graves2012long}
A.~Graves, ``Long short-term memory,'' \emph{Supervised sequence labelling with
  recurrent neural networks}, pp. 37--45, 2012.

\bibitem{chung2014empirical}
J.~Chung, C.~Gulcehre, K.~Cho, and Y.~Bengio, ``Empirical evaluation of gated
  recurrent neural networks on sequence modeling,'' \emph{arXiv preprint
  arXiv:1412.3555}, 2014.

\bibitem{vaswani2017attention}
A.~Vaswani, N.~Shazeer, N.~Parmar, J.~Uszkoreit, L.~Jones, A.~N. Gomez,
  {\L}.~Kaiser, and I.~Polosukhin, ``Attention is all you need,''
  \emph{Advances in neural information processing systems}, vol.~30, 2017.

\bibitem{supratak2017deepsleepnet}
A.~Supratak, H.~Dong, C.~Wu, and Y.~Guo, ``Deepsleepnet: A model for automatic
  sleep stage scoring based on raw single-channel eeg,'' \emph{IEEE
  transactions on neural systems and rehabilitation engineering}, vol.~25,
  no.~11, pp. 1998--2008, 2017.

\bibitem{rajpurkar2017cardiologist}
P.~Rajpurkar, A.~Y. Hannun, M.~Haghpanahi, C.~Bourn, and A.~Y. Ng,
  ``Cardiologist-level arrhythmia detection with convolutional neural
  networks,'' \emph{arXiv preprint arXiv:1707.01836}, 2017.

\bibitem{sharma2022transformer}
G.~Sharma, A.~Dhall, and R.~Subramanian, ``A transformer based approach for
  activity detection,'' in \emph{Proceedings of the 30th ACM International
  Conference on Multimedia}, 2022, pp. 7155--7159.

\bibitem{yu2023semi}
H.~Yu and A.~Sano, ``Semi-supervised learning for wearable-based momentary
  stress detection in the wild,'' \emph{Proceedings of the ACM on Interactive,
  Mobile, Wearable and Ubiquitous Technologies}, vol.~7, no.~2, pp. 1--23,
  2023.

\bibitem{anand2026self}
G.~Anand, A.~M. Kalra, M.~M. Baig, H.~GholamHosseini, E.~Ullah, and W.~Chen,
  ``Self-management of long-term conditions by integrating artificial
  intelligence with wearable devices and internet-of-thing technology: A
  review,'' \emph{Journal of Engineering and Science in Medical Diagnostics and
  Therapy}, vol.~9, no.~1, p. 011106, 2026.

\bibitem{garcia2024sleepecg}
C.~Garc{\'\i}a-Vicente, G.~C. Guti{\'e}rrez-Tobal, F.~Vaquerizo-Villar,
  A.~Mart{\'\i}n-Montero, D.~Gozal, and R.~Hornero, ``Sleepecg-net: explainable
  deep learning approach with ecg for pediatric sleep apnea diagnosis,''
  \emph{IEEE Journal of Biomedical and Health Informatics}, vol.~29, no.~2, pp.
  1021--1034, 2024.

\bibitem{abdelaal2024exploring}
Y.~Abdelaal, M.~Aupetit, A.~Baggag, and D.~Al-Thani, ``Exploring the
  applications of explainability in wearable data analytics: Systematic
  literature review,'' \emph{Journal of Medical Internet Research}, vol.~26, p.
  e53863, 2024.

\bibitem{rossi2023sleep}
M.~Rossi, D.~Sala, D.~Bovio, C.~Salito, G.~Alessandrelli, C.~Lombardi,
  L.~Mainardi, and P.~Cerveri, ``Sleep-see-through: explainable deep learning
  for sleep event detection and quantification from wearable somnography,''
  \emph{IEEE journal of biomedical and health informatics}, vol.~27, no.~7, pp.
  3129--3140, 2023.

\bibitem{li2019enhancing}
S.~Li, X.~Jin, Y.~Xuan, X.~Zhou, W.~Chen, Y.-X. Wang, and X.~Yan, ``Enhancing
  the locality and breaking the memory bottleneck of transformer on time series
  forecasting,'' \emph{Advances in neural information processing systems},
  vol.~32, 2019.

\bibitem{phan2022sleeptransformer}
H.~Phan, K.~Mikkelsen, O.~Y. Ch{\'e}n, P.~Koch, A.~Mertins, and M.~De~Vos,
  ``Sleeptransformer: Automatic sleep staging with interpretability and
  uncertainty quantification,'' \emph{IEEE Transactions on Biomedical
  Engineering}, vol.~69, no.~8, pp. 2456--2467, 2022.

\bibitem{selvaraju2017grad}
R.~R. Selvaraju, M.~Cogswell, A.~Das, R.~Vedantam, D.~Parikh, and D.~Batra,
  ``Grad-cam: Visual explanations from deep networks via gradient-based
  localization,'' in \emph{Proceedings of the IEEE international conference on
  computer vision}, 2017, pp. 618--626.

\bibitem{lundberg2017unified}
S.~M. Lundberg and S.-I. Lee, ``A unified approach to interpreting model
  predictions,'' \emph{Advances in neural information processing systems},
  vol.~30, 2017.

\bibitem{ribeiro2016should}
M.~T. Ribeiro, S.~Singh, and C.~Guestrin, ``" why should i trust you?"
  explaining the predictions of any classifier,'' in \emph{Proceedings of the
  22nd ACM SIGKDD international conference on knowledge discovery and data
  mining}, 2016, pp. 1135--1144.

\bibitem{jin2024cardiologist}
Y.~Jin, Z.~Li, M.~Wang, J.~Liu, Y.~Tian, Y.~Liu, X.~Wei, L.~Zhao, and C.~Liu,
  ``Cardiologist-level interpretable knowledge-fused deep neural network for
  automatic arrhythmia diagnosis,'' \emph{Communications medicine}, vol.~4,
  no.~1, p.~31, 2024.

\bibitem{barkan2021grad}
O.~Barkan, E.~Hauon, A.~Caciularu, O.~Katz, I.~Malkiel, O.~Armstrong, and
  N.~Koenigstein, ``Grad-sam: Explaining transformers via gradient
  self-attention maps,'' in \emph{Proceedings of the 30th ACM International
  Conference on Information \& Knowledge Management}, 2021, pp. 2882--2887.

\bibitem{climeforts}
T.~Ito, K.~Ochiai, and Y.~Fukazawa, ``C-lime: A consistency-oriented lime for
  time-series health-risk predictions,'' in \emph{Knowledge Management and
  Acquisition for Intelligent Systems}, H.~Uehara, T.~Yamaguchi, and Q.~Bai,
  Eds.\hskip 1em plus 0.5em minus 0.4em\relax Cham: Springer International
  Publishing, 2021, pp. 58--69.

\bibitem{Dodge_2025}
\BIBentryALTinterwordspacing
L.~Dodge, ``\BIBforeignlanguage{en-GB}{Geneactiv: Raw data wearable for
  clinical \& public health research},'' Apr. 2025. [Online]. Available:
  \url{https://activinsights.com/digital-health-technologies/professional-wearables/geneactiv/}
\BIBentrySTDinterwordspacing

\bibitem{hfz_category}
\BIBentryALTinterwordspacing
T.~C. Institute, \emph{FITNESSGRAM® Healthy Fitness Zone Performance
  Standards}.\hskip 1em plus 0.5em minus 0.4em\relax California Department of
  Education, jan 2018. [Online]. Available:
  \url{https://resources.finalsite.net/images/v1680962646/lbusdk12caus/agqlptdn6dksijblxbmb/FITNESSGRAM-Healthy-Fitness-Zone-Performance-Standard.pdf}
\BIBentrySTDinterwordspacing

\bibitem{EmpaticaE4}
\BIBentryALTinterwordspacing
E.~Inc., ``Empatica e4 wristband for research,'' 2025. [Online]. Available:
  \url{https://www.empatica.com/en-int/research/e4/}
\BIBentrySTDinterwordspacing

\bibitem{american1999sleep}
A.~A. of~Sleep~Medicine \emph{et~al.}, ``Sleep-related breathing disorders in
  adults: recommendations for syndrome definition and measurement techniques in
  clinical research. the report of an american academy of sleep medicine task
  force,'' \emph{Sleep}, vol.~22, pp. 667--689, 1999.

\bibitem{Phillips2013}
L.~R.~S. Phillips, G.~Parfitt, and A.~V. Rowlands, ``Calibration of the genea
  accelerometer for assessment of physical activity intensity in children,''
  \emph{Journal of Science and Medicine in Sport}, vol.~16, no.~2, pp.
  124--128, 2013.

\bibitem{phillips2013calibration}
L.~R. Phillips, G.~Parfitt, and A.~V. Rowlands, ``Calibration of the genea
  accelerometer for assessment of physical activity intensity in children,''
  \emph{Journal of science and medicine in sport}, vol.~16, no.~2, pp.
  124--128, 2013.

\bibitem{liu2019variance}
L.~Liu, H.~Jiang, P.~He, W.~Chen, X.~Liu, J.~Gao, and J.~Han, ``On the variance
  of the adaptive learning rate and beyond,'' \emph{arXiv preprint
  arXiv:1908.03265}, 2019.

\bibitem{ekelund2019dose}
U.~Ekelund, J.~Tarp, J.~Steene-Johannessen, B.~H. Hansen, B.~Jefferis, M.~W.
  Fagerland, P.~Whincup, K.~M. Diaz, S.~P. Hooker, A.~Chernofsky \emph{et~al.},
  ``Dose-response associations between accelerometry measured physical activity
  and sedentary time and all cause mortality: systematic review and harmonised
  meta-analysis,'' \emph{bmj}, vol. 366, 2019.

\bibitem{uddin2020clustering}
R.~Uddin, E.-Y. Lee, S.~R. Khan, M.~S. Tremblay, and A.~Khan, ``Clustering of
  lifestyle risk factors for non-communicable diseases in 304,779 adolescents
  from 89 countries: A global perspective,'' \emph{Preventive medicine}, vol.
  131, p. 105955, 2020.

\bibitem{jiang2023healthprism}
Z.~Jiang, H.~Chen, R.~Zhou, J.~Deng, X.~Zhang, R.~Zhao, C.~Xie, Y.~Wang, and
  E.~C. Ngai, ``Healthprism: a visual analytics system for exploring children's
  physical and mental health profiles with multimodal data,'' \emph{IEEE
  Transactions on Visualization and Computer Graphics}, vol.~30, no.~1, pp.
  1205--1215, 2023.

\end{thebibliography}
\end{document}